\documentclass[runningheads]{llncs}
\usepackage{graphicx}

\usepackage{tikz}
\usepackage{comment}
\usepackage{amsmath,amssymb} 
\usepackage{color}
\usepackage[accsupp]{axessibility}  

\usepackage{nicefrac}
\usepackage{wrapfig}

\usepackage{bm}
\usepackage{mathrsfs}
\usepackage{multirow}
\usepackage{caption}
\usepackage{array}
\usepackage{pifont}
\usepackage{color,xcolor}
\usepackage{algorithm, algpseudocode}
\definecolor{mydarkblue}{rgb}{0,0.08,0.45}

\begin{document}
\pagestyle{headings}
\mainmatter

\title{
Controllable Evaluation and Generation of Physical Adversarial Patch on Face Recognition} 

\titlerunning{Evaluation and Generation of Physical Adversarial Patch} 
\author{Anonymous ECCV submission}
\institute{Paper ID \ECCVSubNumber}

%
\author{\small Xiao Yang$^{1}$, Yinpeng Dong$^{1,2}$, Tianyu Pang$^{1}$, Zihao Xiao$^{2}$, Hang Su$^{1}$, Jun Zhu$^{1,2}$ }
\authorrunning{X.Yang et al.}
%
\institute{$^{1}$ Department of Computer Science \& Technology, Tsinghua University \\
$^{2}$ RealAI \\
\email{\tt\scriptsize \{yangxiao19, dyp17, pty17\}@mails.tsinghua.edu.cn \hspace{2ex} zihao.xiao@realai.ai  \\
	\{suhangss,dcszj\}@tsinghua.edu.cn}}
\maketitle

\begin{abstract}
Recent studies have revealed the vulnerability of face recognition models against physical adversarial patches, which raises security concerns about the deployed face recognition systems.
However, it is still challenging to ensure the reproducibility for most attack algorithms under complex physical conditions, which leads to the lack of a systematic evaluation of the existing methods. It is therefore imperative to develop a framework that can enable a comprehensive evaluation of the vulnerability of face recognition in the physical world. To this end, we propose to simulate the complex transformations of faces in the physical world via 3D-face modeling, which serves as a digital counterpart of physical faces. The generic framework allows us to control different face variations and physical conditions to conduct reproducible evaluations comprehensively. With this digital simulator, we further propose a \textbf{Face3DAdv} method considering the 3D face transformations and realistic physical variations. Extensive experiments validate that Face3DAdv can significantly improve the effectiveness of diverse physically realizable adversarial patches in both simulated and physical environments, against various white-box and black-box face recognition models.

\end{abstract}

\vspace{-3ex}
\section{Introduction}
\label{sec:introduction}

Face recognition, as a prevailing task in computer vision, has experienced substantial improvements thanks to the rapid development of deep neural networks (DNNs)~\cite{deng2018arcface,wen2016discriminative}. DNNs facilitate the broad application of face recognition in various safety-critical fields, including finance/payment, public access, surveillance, etc.
However, face recognition models based on DNNs are vulnerable to \emph{adversarial examples}~\cite{sharif2016accessorize,dong2019efficient,yang2020delving,komkov2021advhat,xiao2021improving,tong2021facesec} 
--- maliciously generated inputs
to mislead a target model, which may lead to serious consequences or security problems in real-world applications.

\begin{figure}[t]
\begin{center}
\includegraphics[width=0.9\linewidth]{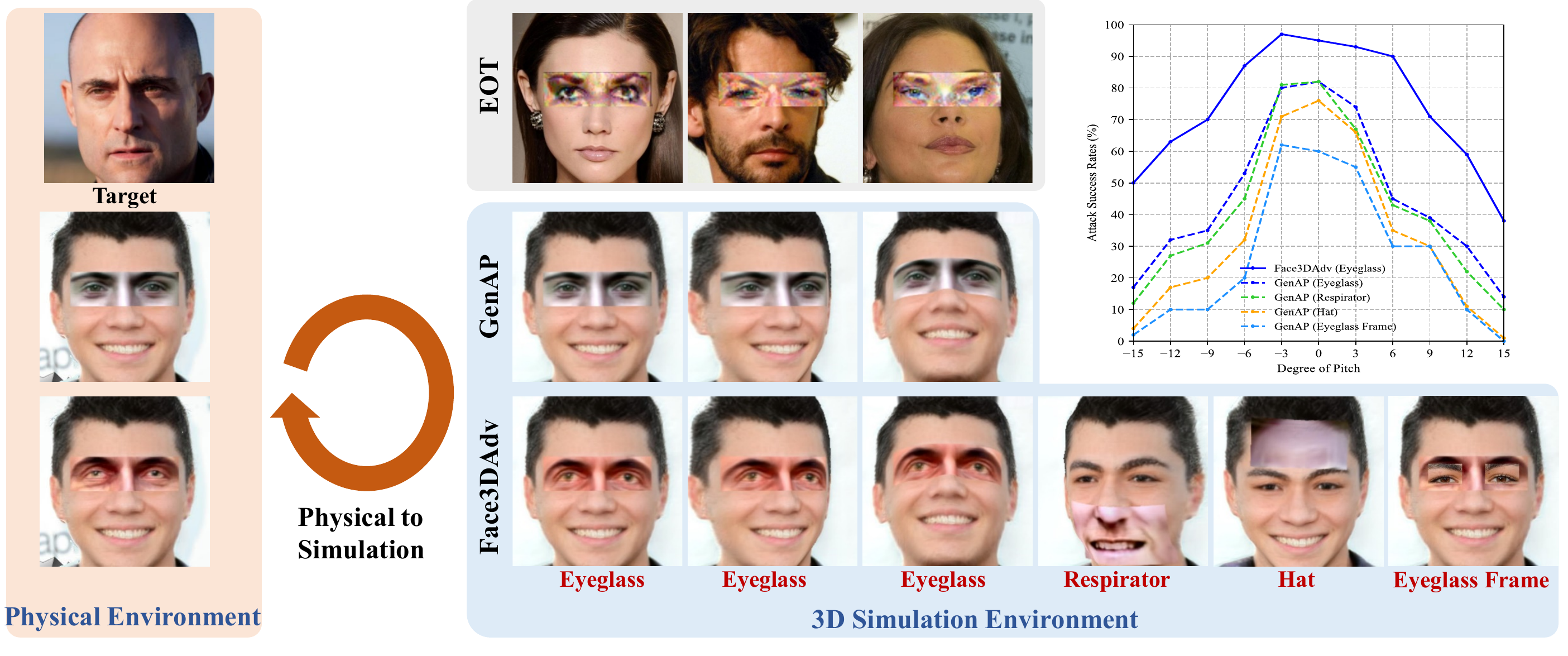}
\end{center}
\vspace{-4ex}
\caption{The adversarial patches crafted by EOT~\cite{Athalye2017Synthesizing} with a typical pasting method, and GenAP~\cite{xiao2021improving} and our Face3DAdv in the simulated environment for impersonation attack. The examples are derived from different physically realizable attacks, including Eyeglass~\cite{xiao2021improving}, Respirator~\cite{tong2021facesec}, Hat~\cite{komkov2021advhat}, and Eyeglass frame~\cite{sharif2016accessorize,sharif2017adversarial}, which have verified their practicality in many \emph{unattended} recognition scenarios. We also show the results of attack success rate against ArcFace~\cite{deng2019arcface} over $[-15,15]$ degrees of the pose of \emph{pitch} under a certain degree of \emph{yaw}, demonstrating consistent effectiveness of our method. More results regarding effectiveness and practicality are presented in Sec.~\ref{sec:exp}.}
\label{fig:intro}
\vspace{-2ex}
\end{figure}

Extensive efforts have been devoted to studying the generation of adversarial examples (i.e., adversarial attacks) on face recognition models, which can be conducive to investigating model robustness~\cite{yang2020delving,tong2021facesec}.
Some work~\cite{sharif2016accessorize,yang2020delving,dong2019efficient,yang2021towards} has proposed to apply minimal perturbations (measured by the $\ell_p$ norm) to face images in the \emph{digital} world, aiming to evade being recognized or to impersonate another identity. 
However, practical face recognition systems usually process face photos taken in the \emph{physical} world. 
Thus, it is of particular importance to explore physical adversarial attacks to identify the weaknesses of these models before they are deployed. 
To this end, some typical approaches generate various adversarial patches \cite{brown2017adversarial} that are wearable on faces, including eyeglass frames \cite{sharif2016accessorize,sharif2017adversarial,xiao2021improving}, hats~\cite{komkov2021advhat}, and stickers~\cite{guo2021meaningful,shen2021effective}. Although the patches are noticeable to human eyes, the attack mechanism can still take effect in automatic face recognition systems in many \emph{unattended} scenarios, e.g., deceiving the payment system in vending machines~\cite{geekpwn2020} and unlocking a mobile phone or car~\cite{vee2019}.




Despite the success, the existing physical attack methods on face recognition still have several limitations.
First, there is no systematic testing protocol for physical attacks. The evaluation is usually conducted
by asking a handful of volunteers to attach the adversarial patches, followed by testing in a specific environment (e.g., printer, viewpoints, lighting conditions)~\cite{sharif2016accessorize,komkov2021advhat,guo2021meaningful}, making it hard to evaluate and compare the effectiveness of different methods. The 
inconsistent experimental settings also limit the reproducibility of physical adversarial examples in different conditions. 
Second, most methods aim to craft adversarial examples robust to varying physical conditions by optimizing over 2D image transformations~\cite{Athalye2017Synthesizing}, such as rotation, translation and additive Gaussian noise,
but they fail to consider other physical variations of 3D faces, such as viewpoint and lighting.
Third, the printed adversarial patch needs to be placed on relatively flat face areas, including eyeglass
frames~\cite{sharif2016accessorize}, hats~\cite{komkov2021advhat}, or respirators~\cite{xiao2021improving}. A few studies considered simple geometric transformations of the patch (e.g., parabolic~\cite{komkov2021advhat}), resulting in inferior performance when fitting the patch to the real 3D face due to the inevitable deformation.


In this paper, we propose a novel \emph{simulation framework} which can reflect the characteristics of physical faces, enabling us to conduct fair and comprehensive evaluations of physical attacks on face recognition. The accessible and general framework pairs attackers acting in a simulated environment with counterparts acting in a realistic physical environment, 
which can serve as a benchmarking tool to produce the replicable research. 
To make the physical conditions easily controllable and the synthesized face images photorealistic under different conditions in simulation, we leverage the recent advances in 3D face modeling to build the simulator.
Specifically, we adopt a 3D generator \cite{shi2021lifting} to synthesize 3D face information, including texture, shape, viewpoint, and lighting, using only a single-view face image. Then, we propose a texture-based adversarial attack paradigm to generate a 3D adversarial face, which can naturally stitch a patch onto the face to make the adversarial patch more versatile and realistic. Finally, after introducing a differentiable renderer~\cite{ravi2020accelerating}, we can obtain 2D adversarial faces under diverse physical variations. 
Once informed with such a simulator, we can use the virtual environment to study the robustness of different models on face recognition, providing us a possibility to make improvements to the models especially in the physical world, as shown in Fig.~\ref{fig:intro}.



Based on this simulation framework, the attacker has the ability to develop more reliable physical adversarial attacks by controlling the simulated environments, thus enabling the crafted adversarial patches to be more robust to physical transformations. To demonstrate this, we propose a \textbf{Face3DAdv} attack method to generate robust adversarial patches by exploiting physical transformations in adversarial scenarios based on the simulation framework. Moreover, since the physical variations are much more abundant in our method, we adopt a more effective strategy to focus on favorable transformations within a principled optimization framework. As a comparison, the previous methods~\cite{Athalye2017Synthesizing,sharif2016accessorize} typically select physical transformations fully at random to optimize robust perturbations, without considering the different importance of physical variations. Extensive experiments demonstrate that our Face3DAdv achieves consistent improvements in both simulated and physical environments. Besides, 3D adversarial patches crafted by Face3DAdv is also more conducive to steadily passing defensive mechanism (\emph{commercial Live Detection API}) in automatic face recognition systems.


To the best of our knowledge, this is the first attempt that conducts a reproducible physical-world adversarial attacks on face recognition, especially including 3D face recognition models. Our contributions can be summarized
as 
\begin{itemize}
    
    \vspace{-1ex}
    \item We construct a comprehensive evaluation protocol for facilitating the fair and convenient evaluation of physical attacks on face recognition by deriving a simulation framework, which can well approximate the attack performance in the physical world.

    \item We propose a 3D-aware attack method --- Face3DAdv to generate robust adversarial patches, showing significant improvements over the previous methods with a particular focus on diverse physical conditions of 3D transformations, lighting variations, etc. 
    
    \item We provide a novel framework to explore the vulnerabilities of face recognition models, which can facilitate the development of more robust models.   
\end{itemize}


\vspace{-1ex}
\section{Related Work}

\textbf{Adversarial attacks in the physical world.}
Recent work has shown that adversarial examples~\cite{biggio2013evasion,goodfellow2014explaining,szegedy2013intriguing} can exist in the physical world~\cite{Kurakin2016,Athalye2017Synthesizing}, resulting in an emerging threat.
In particular, adversarial patches~\cite{brown2017adversarial} only perturb a small cluster of pixels, and can be applied to real objects in the physical world~\cite{Eykholt_2018_CVPR,zhao2020defenses,xu2020adversarial,zolfi2021translucent,yang2020design}.
Adversarial patches on face recognition have also been explored~\cite{sharif2016accessorize,brown2017adversarial}. 
By attaching a carefully generated patch to the face, some studies~\cite{pautov2019adversarial,komkov2021advhat} have shown success of physical attacks against the state-of-the-art face recognition models. However, these methods did not consider the face variations in the physical world, thus resulting in performance degeneration in real testing scenarios. Meanwhile, existing physical attacks commonly use EOT~\cite{Athalye2017Synthesizing} by randomly sampling the transformations during optimization without considering the different importance for the diverse physical variations. 


\textbf{3D face modeling.}
As one of the popular 3D face modeling mechanisms, 3D Morphable Model (3DMM) is commonly adopted to represent faces~\cite{tuan2017regressing}, which are parameterized by identity, expression, and illumination. Although 3DMM offers control over the semantic parameters, it suffers from photorealism and models only the essential parts of 
a portrait image (e.g., hair, mouth interior, background). More recent work reconstructs plausible 3D face shapes by exploiting knowledgeable parameter metrics of 3DMM~\cite{tewari2020stylerig,deng2020disentangled}. On the other hand, some face representation methods leverage 3D position maps~\cite{shi2021lifting,henderson2020leveraging} to represent and output the mesh of the target, and achieve the controllable parametric nature of existing face models. Therefore, we can construct a flexible environment that simulates the physical world with the aid of these blossoming techniques on 3D face modeling.  The digital surrogate of a real face provides us a possible solution to conduct reliable and reproducible evaluation for facilitating physical attacks. 

\vspace{-1ex}
\section{Robust Evaluation for Physical Attacks}


The current physical attacks on face recognition~\cite{sharif2016accessorize,guo2021meaningful,komkov2021advhat} are usually evaluated by: 1) printing adversarial patches (e.g., eyeglass frames, hats, etc); 2) asking a few volunteers to attach them; and 3) testing the attack performance under a specific environment. However, the evaluation methodology is insufficient due to the lack of a systematic testing protocol. The experimental settings (including printer, chosen volunteers, and physical environment) are obviously inconsistent across different research, making it hard to compare and evaluate the 
effectiveness of existing methods. Such inconsistency significantly limits the reproducibility of physical attacks, hindering their further development.

\begin{figure*}[t]
\begin{center}
\includegraphics[width=0.95\linewidth]{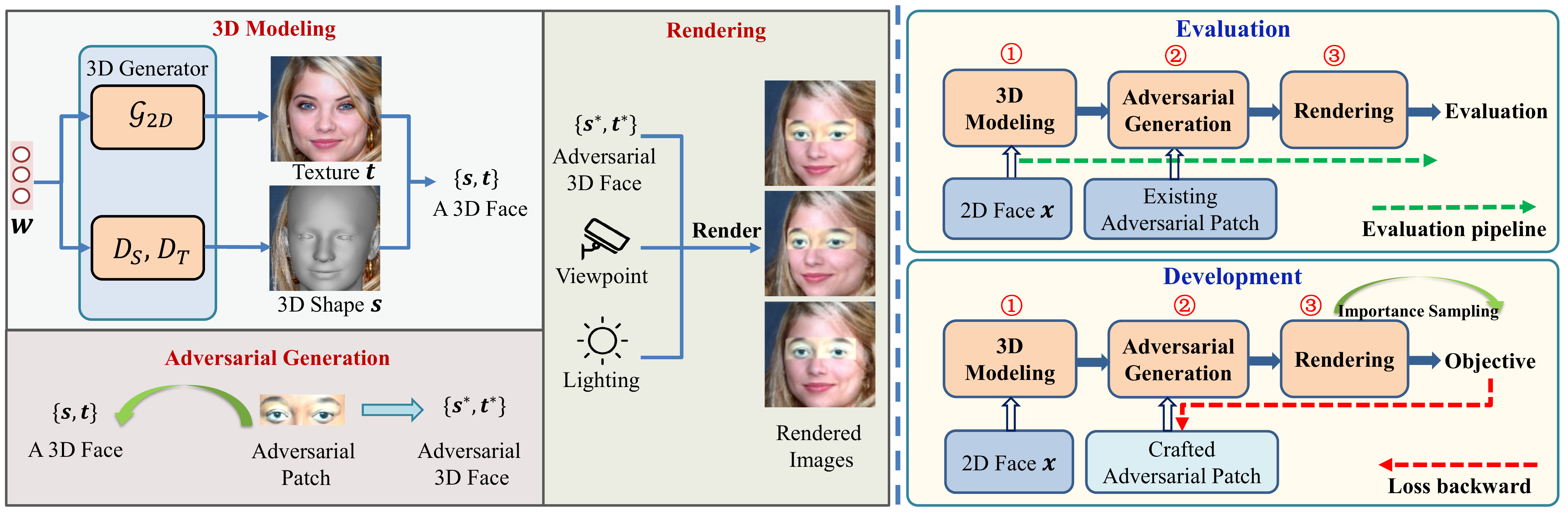}
\end{center}
\vspace{-3.5ex}
\caption{The overall simulation framework for evaluating and developing physical attacks include four modules of \textbf{3D Modeling}, \textbf{Adversarial Generation}, \textbf{Rendering}, and \textbf{Evaluation}. \textbf{3D Modeling} reconstructs a 3D face $\{\bm{s},\bm{t}\}$ that can be manipulated, including a 2D texture face $\bm{t}$ from the generative model $\mathcal{G}_{2D}$ and a shape representation $\bm{s}$ from deep networks of shape $D_S$ and transformation map $D_T$.  \textbf{Adversarial Generation} adopts a texture-based adversarial method to apply a patch to a certain region to generate an adversarial 3D face $\{\bm{s}^*,\bm{t}^*\}$. \textbf{Rendering} adopts a renderer to produce a series of 2D rendered adversarial faces given $\{\bm{s}^*,\bm{t}^*\}$, different viewpoints and lighting conditions. We also present the evaluation phase that aims to evaluate the performance of the existing attacks, and the development phase that utilizes diverse physical conditions to boost the performance of adversarial attacks.}
\label{fig:framework}
\vspace{-2ex}
\end{figure*}

To address this problem, it is imperative to develop a reproducible  framework for readily and fairly evaluating the performance of physical adversarial attacks on face recognition. We advocate using a simulator rather than performing experiments in the physical world for the following reasons: 1) \textbf{completeness:} the simulator can provide a complete picture of the effectiveness of different attack methods given various controllable physical conditions; 2) \textbf{fair and replicable comparisons:} based on the same simulation framework, the comparisons between different attacks are fairer, and the evaluation results are replicable; 3) \textbf{cheap and easy to support large experiments:} conducting experiments on the simulation framework is much cheaper and easier, which supports larger-scale experiments, rather than only inviting a few volunteers to evaluate the performance in the physical world. 

However, it is non-trivial to build a simulation framework that can be reliably used as a testbed for physical adversarial attacks on face recognition.
It should at least satisfy the following requirements. First, most common physical conditions, including 2D image transformations, viewpoint, and lighting, need to be controlled in a fine-grained manner, such that we can measure the attack performance under each condition. Second, the simulator must be able to synthesize photorealistic face images under different physical conditions, e.g., without losing the identity information. Third, the performance 
evaluated by the simulator should be consistent with 
that in the physical world. In this paper, we present a simulation framework based on 3D face modeling to fulfill the first two requirements, and empirically verify that our simulator can well approximate the performance of adversarial attacks in the physical world.

As illustrated in Fig.~\ref{fig:framework}, our simulation framework consists of four modules: 3D face modeling, adversarial generation, rendering, and evaluation. We introduce the details of these four steps in the following discussion.

\textbf{3D face modeling.} First, we try to reconstruct a 3D face representation that can be easily manipulated. The 3D face is expected to approximate a real face in the physical world. 
Given a 2D face image $\bm{x}\in\mathcal{X}$, we can exploit 3D parametric fitting to generate a 3D face by leveraging the recent advances in 3D face modeling. 
In this paper, we adopt the state-of-the-art pre-trained 3D generator of $\mathcal{G}_{3D}$~\cite{shi2021lifting} for 3D face modeling, which can disentangle the generation process of a 2D generator $\mathcal{G}_{2D}$ instantiated by StyleGAN~\cite{Karras2019A} into different 3D modules for a 3D shape representation.
Therefore, we can obtain a 3D face representation from a style code $\bm{w}$ including a 3D shape representation of $\bm{s}$ and a 2D texture face of $\bm{t}$ from $\mathcal{G}_{2D}$.

Given a face recognition model $f(\bm{x}):\mathcal{X}\rightarrow\mathbb{R}^d$,
we optimize the parameter of $\bm{w}$ for the generator 
by minimizing the distance between the original face image and the rendered image of ${\bm{x}'}$ as 
\begin{equation}
    \label{eq:object}
    \min_{\bm{w}} \mathcal{D}_{f}(\bm{x}', \bm{x}) + \lambda\|\bm{x}'-\bm{x}\|_{1},
\end{equation}
where ${\bm{x}'} :=  \mathrm{R}(\mathcal{G}_{3D}(\bm{w}); V_{0}, L_{0})$ with $\mathrm{R}$ being a differentiable renderer, and $V_{0}$ and $L_{0}$ are corresponding parameters of neutralized viewpoint and lighting; 
and $\lambda$ is a balancing hyperparameter. We adopt the $\ell_1$ norm in the objective since the $\ell_2$ norm can lead to blurry textures~\cite{huh2020transforming}. $\mathcal{D}_{f}$ computes the distance of the feature representations of $f$ as
\begin{equation}
    \mathcal{D}_f(\bm{x}', \bm{x}) = \|f(\bm{x}') - f(\bm{x})\|_2^2.
\end{equation}
By optimizing the objective function~(\ref{eq:object}), we can obtain the optimal $\bm{w}^{*}$ and get the 3D face as $\{\bm{s}, \bm{t}\} = \mathcal{G}_{3D}(\bm{w}^{*})$.


\textbf{Adversarial generation.} 
The next step is to apply the adversarial examples to the 3D face model. The existing attack methods usually adopt texture-based adversarial patches, i.e., for a face image $\bm{x}$, these methods can generate an adversarial face image $\bm{x}^*$ by applying a patch to a certain region. Since they do not modify the face shape, we directly replace the texture $\bm{t}$ of the original face image $\bm{x}$ as $\bm{t}^*=\bm{x}^*$. Notably, our framework can also perform adversarial attacks on face shapes if necessary for new attackers.

\textbf{Rendering 2D images with transformations.} Given a 3D adversarial face $\{\bm{s}, \bm{t}^*\}$, we can adopt a renderer~\cite{ravi2020accelerating} to produce 2D rendered adversarial faces given different viewpoints and lighting conditions. Specifically, we choose a set of viewpoints $V = \{V_i\}_{i=1}^{N_v}$ and lighting variations $L = \{L_j\}_{j=1}^{N_l}$, and then render an adversarial image as
\begin{equation}
\label{eq:base}
 \bm{r}^{*}_{(i,j)}= \mathrm{R}(\bm{s}, \bm{t}^*; V_{i}, L_{j}).
\end{equation}
We can also apply some 2D image transformations (e.g., rotation, translation, scaling, etc) to $\bm{r}^{*}_{(i,j)}$.

\textbf{Evaluation.} The final step is to evaluate the performance of the attack by feeding the rendered adversarial images into the face recognition model. For different tasks, we can evaluate the attack performance in different ways.

\textbf{Discussion on the printer.}
Some previous research~\cite{sharif2016accessorize,thys2019fooling,xu2020adversarial,zheng2021robust} has studied the color deviation between the digital image and its printed version by mapping a digital color spectrum to printed counterpart or adopting non-printability losses. However, we mainly focus on constructing a simulation framework for evaluating physical adversarial attacks. The previous approaches are generally compatible with our framework in the physically realizable procedure. 


\vspace{-1ex}
\section{Face3DAdv}
\vspace{-1ex}

In this section, we propose a \textbf{Face3DAdv} attack method to exploit the various physical transformations.

\vspace{-2ex}
\subsection{Preliminary}
\vspace{-1ex}

Face recognition usually has two sub-tasks: face verification and
face identification~\cite{huang2008labeled}.
We mainly consider face verification in this paper, while the proposed approach can be naturally extended to face identification.
In face verification, the feature distance between a pair of images $\{\bm{x}^{a}, \bm{x}^{b}\} \subset \mathcal{X}$ is first calculated as $\mathcal{D}_f(\bm{x}^a, \bm{x}^b)$.
Then the prediction of face verification can be formulated as
\begin{equation}\label{eq:delta}
    \mathcal{C} (\bm{x}^{a}, \bm{x}^{b}) = \mathbb{I} (\mathcal{D}_f (\bm{x}^{a}, \bm{x}^{b}) < \delta ),
\end{equation}
where $\mathbb{I}$ is the indicator function, and $\delta$ is a threshold. When $\mathcal{C}(\bm{x}^{a}, \bm{x}^{b})=1$, the two images are recognized as the same identity, otherwise different identities. Note that this definition is consistent with the commonly used cosine similarity metric, since $f$ outputs a normalized feature.

Given the original face images $\bm{x}^{a}$ and $\bm{x}^{b}$, we aim to generate an adversarial image $\bm{x}^*$ by adding a perturbation to $\bm{x}^{a}$ to mislead the face recognition model when recognizing $\bm{x}^*$ and $\bm{x}^{b}$.
There are generally two types of adversarial attacks on face recognition: \emph{dodging} and \emph{impersonation}.
A dodging attack aims to make the face recognition model fail to recognize the identity of $\bm{x}^*$, i.e., to make 
$\mathcal{C} (\bm{x}^{*}, \bm{x}^{b}) = 0$ while $\mathcal{C} (\bm{x}^{a}, \bm{x}^{b}) = 1$; an impersonation attack aims to make the face recognition model recognize $\bm{x}^{*}$ as a specific identity, i.e., 
to make $\mathcal{C} (\bm{x}^{*}, \bm{x}^{b}) = 1$ while $\mathcal{C} (\bm{x}^{a}, \bm{x}^{b}) = 0$.

\vspace{-0.1ex}
\subsection{3D-Aware Adversarial Attack}
\vspace{-0.1ex}

To facilitate the physical realizability of the adversarial examples, we study adversarial patches that are restricted to a specifically designed region. Although some elaborate adversarial patches~\cite{komkov2021advhat,xiao2021improving} consider 2D image transformations, they do not take into account other realistic 3D physical transformations, thus leading to inevitable degeneration of their effectiveness. To make the crafted adversarial patch more versatile and effective in the real world, we optimize the adversarial patch over both the common 2D transformations and the newly considered 3D transformations. 
Based on our simulation framework, we can readily optimize the adversarial patches over 3D transformations.
Therefore, the attack objective function of crafting adversarial examples can be formulated as
\begin{small}
\begin{gather}
\label{eq:mask}
    \min_{\bm{s}^{*}, \bm{t}^{*}}\; \mathbb{E}_{V_{i} \sim V, L_{j} \sim L} [\mathcal{J}_{f}(\mathrm{R}(\bm{s}^{*}, \bm{t}^{*}; V_{i}, L_{j}), \bm{x}^{b})],  \\
    \text{s.t.} (\bm{1} - \mathrm{M})\odot\mathrm{R}(\bm{s}^{*}, \bm{t}^{*}; V_{i}, L_{j}) = (\bm{1} - \mathrm{M})\odot\mathrm{R}(\bm{s}^{a}, \bm{t}^{a}; V_{i}, L_{j}), \nonumber\end{gather}\end{small}where $M\in\{0,1\}^n$ is a binary mask to apply the perturbations to pixels where the value of the mask is $1$, $\odot$ is the element-wise multiplication operation, $\{\bm{s}^a,\bm{t}^{a}\}$ is the 3D face obtained by optimizing Eq.~\eqref{eq:object} given a 2D face image $\bm{x}^{a}$ and $\mathcal{J}_{f}$ is the attack loss. In this paper, we adopt $\mathcal{J}_{f} = -\mathcal{D}_{f}$ for a dodging attack and $\mathcal{J}_{f} = \mathcal{D}_{f}$ for an impersonation attack. Since 2D transformations~\cite{xie2019improving} are generally compatible with the objective~(\ref{eq:mask}), we can craft a 3D adversarial face to fool the face recognition systems for diverse 2D and 3D face transformations. 

\textbf{Mapping shape representation.} Note that the optimization problem (\ref{eq:mask}) is constrained, which must ensure that the shape representation $\bm{s}^{*}$ of the 3D adversarial face is only modified in the designed region in every optimization step. However, this can give rise to the inconsistency of $\bm{s}^{*}$  in the designed mask region and original face representation $\bm{s}^{a}$ after a long optimization trajectory, consequently leading to inevitable performance degradation due to shape disharmony of the whole 3D face. To address this issue, we directly reduce the optimization space of the 3D adversarial face by replacing $\bm{s}^{*}$ with $\bm{s}^{a}$ in the optimization. In this way, we can entirely restrict the 3D adversarial face in a prior fixed shape, and only optimize the texture map $\bm{t}^{*}$.

Given the objective function in Eq.~\eqref{eq:mask}, we can iteratively apply the fast gradient method~\cite{Kurakin2016} with a small step size $\alpha$ to generate adversarial examples. In particular, we optimize the adversarial texture image $\bm{t}^{*}$ via
\begin{equation}
\label{eq:step}
    \bm{t}_{k+1}^{*}=\Pi_{D^t}\left(\bm{t}_{k}^{*}-\alpha \cdot \mathrm{M}\odot \mathrm{sign}\left(g_{k+1}\right)\right),
\end{equation}
where $g_{k+1}$ is the updated gradient at the $(k+1)$-th iteration, and $\Pi_{D^t}$ is the projection function that projects the adversarial
images onto the $D^t = \{\bm{t}:\|\mathrm{M}\odot\bm{t} - \mathrm{M}\odot\bm{t}^{a}\|_{\infty} \leq \epsilon \}$.
We call it \textbf{Face3DAdv ($\bm{x}$)}. Besides, $\bm{t}^{*}$ can be optimized in the latent space $\bm{w}^{*}$ in $\mathcal{G}_{2D}$ by following a state-of-the-art transferable adversarial method~\cite{xiao2021improving} on face recognition, which can be formulated as $\bm{t}^{*} = \mathcal{G}_{2D}(\bm{w}^{*})$.  And $\bm{w}^{*}$ can be optimized by adopting a popular optimizer, such as Adam~\cite{Kingma2014}, which is called \textbf{Face3DAdv ($\bm{w}$)}.

\subsection{Optimization by Importance Sampling} 
\label{sec:is}

The typical EOT~\cite{Athalye2017Synthesizing} randomly selects transformations to craft adversarial examples during optimization, without considering the importance among different transformations. As illustrated in Fig.~\ref{fig:intro}, we show the heatmaps of impersonation attacks under different face variations, which motivates us to conduct a more effective sampling strategy to learn the more difficult or critical transformations. 

\begin{algorithm}[t]\small
    \caption{Face3DAdv}\label{algo1}
\begin{algorithmic}[1]
\Require A pre-trained 3D generative model $\mathcal{G}_{3D}$, a FR model ${f}$, a real face image $\bm{x}^{a}$, a target face image $\bm{x}^{b}$, 2D transformation function $T$.
\Ensure Adversarial image $t^*$. 

\For{iter in MaxIterations $N_1$} {\color{blue}{\Comment{\emph{Stage I: Obtain a 3D face}}}}
\State Initialize latent code $\bm{w} = \bm{w}_{0}$;
\State         Obtain $\mathcal{J}$ from Eq.~\eqref{eq:object};
\State $\bm{w} \leftarrow \bm{w}-\eta \nabla_{w} \mathcal{J}$;
\EndFor

\State  Forward pass the optimal $\bm{w}^{*}$ into $\mathcal{G}_{3D}$ to the 3D face $\{\bm{s}^a,\bm{t}^a\}$; 

\State Initializing $\bm{t}^{*}_{0} = \bm{x}^{b}$; {\color{blue}{\Comment{\emph{Stage II: Optimize $t^*$}}}}
\For{$k$ in MaxIterations $N_2$}
\State $\bm{t}_{k}^{*} = \bm{t}^{a} \odot (\bm{1} - \mathrm{M}) +  {\bm{t}}_{k}^{*} \odot \mathrm{M}$;
\State Construct 3D adversarial face $\{ \bm{s}^{a},\bm{t}_{k}^{*}\}$;
\State Get importance probability $\hat{P}_{i,j}$ from Eq.~\eqref{eq:prob};
\State Draw $M$ rendered images $\{\bm{r}^{*}_{k,m}\}_{m=1}^{M}$ according to $\hat{P}$;
\State Obtain the gradient  $\bm{g}_{k+1} = \nabla_{\bm{t}} \mathcal{J}_{f}(\sum_{m}T_{m}(\bm{r}^{*}_{k,m}), \bm{x}^{b})$;
\State Update $\bm{t}^{*}_{k+1}$ via Eq.~\eqref{eq:step};
\EndFor

\end{algorithmic}
\end{algorithm}
        
    
        
        
        

        
    

Given an adversarial patch, a larger loss $\mathcal{J}_f$ on the condition $\{V_{i},L_{j}\}$ represents a greater attack difficulty. Thus, we can utilize $\mathcal{J}_f$ as a surrogate for evaluating the usefulness of the condition $\{V_{i},L_{j}\}$. Those transformations with larger losses should be selected more frequent in the optimization phase, yielding a more effective learning strategy.

To achieve this, we define a flexible importance sampling strategy in every iteration through a probability distribution $P$, where $P_{i,j}$ indicates sampling probability on the condition $\{V_{i},L_{j}\}$, which can be represented by a softmax function based on Eq.~\eqref{eq:mask} as
\begin{equation}
\label{eq:prob}
    P_{i,j} = \frac{1}{Z}e^{\mathcal{J}_{f}(\mathrm{R}(\bm{s}^{a},\bm{t}^{*};V_{i}, L_{j}), \bm{x}^{b})}, 
\end{equation}
where $Z={\sum_{i,j}e^{\mathcal{J}_{f}(\mathrm{R}(\bm{s}^{a},\bm{t}^{*};V_{i}, L_{j}), \bm{x}^{b})}}$ is the normalization factor. 
Therefore, if a loss value is larger, we assign a larger value to $P_{i,j}$ such that the transformation $\{V_{i}, L_{j}\}$ will be selected with higher probabilities in the optimization trajectory. In each iteration, we sample the points with a batch size of $k$ according to $P$. 
The detailed optimization procedure of \textbf{Face3DAdv ($\bm{x}$)} is summarized in Algorithm~\ref{algo1}, which can be easily extended to \textbf{Face3DAdv ($\bm{w}$)} by optimizing the latent code $\bm{w}^*$ for obtaining $\bm{t}^*$ in Stage II. 

\vspace{-3ex}
\section{Experiments}\label{sec:exp}
\vspace{-1ex}

In this section, we first present a simulation-based evaluation framework, and present the experimental results to demonstrate the effectiveness of our proposed Face3DAdv. Finally, we also validate that the simulator can well approximate the performance of attacks in the physical world.

\vspace{-2ex}
\subsection{Experimental Settings}

\textbf{Testing protocol.} To facilitate the fair and convenient evaluation of physical attacks on face recognition, we aim to construct a comprehensive testing protocol. Although previous methods~\cite{zheng2021robust} have considered the  significance of evaluating physical variations, e.g., certain poses and lighting in practical scenarios, it is still difficult to conduct a fair comparison between different methods due to poor reproducibility. To tackle this problem, we first customize realistic transformation conditions based on the simulation framework to reduce the potential bias by an uncontrolled experimenter. In our simulation framework, we conduct a total of 200 experimenters from LFW~\cite{huang2008labeled} and CelebA-HQ~\cite{karras2017progressive}, which are two of the most widely used benchmark datasets on both low-quality and high-quality face images. For every experimenter, we introduce a controllable environmental framework, including different poses and lightings, for reliable evaluation of adversarial attacks. As for poses, we choose 3D face variations, i.e., \emph{yaw} and \emph{pitch}. These two variations of all experimental faces are required to have specific movement ranges of the cruciform rail from $-15$ to $15$ angles, respectively. Meanwhile, we also create a series of relighted testing images by creating a shading map of \emph{lighting} from left to right. Furthermore, we have linearly combined these thee conditions to constitute a new type, named \emph{mixture}. As a comparison, previous methods~\cite{xie2019improving,komkov2021advhat} have modeled 2D image transformations, which lack practical consideration in real-world scenarios for adversarial attacks. The detailed testing protocol and results of 2D transformations are provided in Appendix {\color{red} A}.

\textbf{Networks.}
We use three face recognition models for evaluation --- ArcFace~\cite{deng2019arcface}, CosFace~\cite{wang2018cosface}, and FaceNet~\cite{schroff2015facenet}. These models
have different model architectures and training objectives. For each model, we first compute the optimal threshold by following the standard protocol in the LFW dataset, which obtains over 99\% benign recognition accuracy on LFW. If the distance of two images that are fed into the model exceeds the threshold, we regard them as different identities; otherwise, as the same identities.

\begin{table*}[!t]
    \begin{center}
    \scriptsize
    \caption{The attack success rates ($\%$) of the different face recognition models against impersonation attacks on LFW with adversarial glasses. $^*$ indicates white-box attacks.}
    \label{tab:transfer}
    \setlength{\tabcolsep}{0.5pt}
    \begin{tabular}{p{4ex}<{\centering}|c|ccc|ccc|ccc|ccc}
    \hline
         & \multirow{2}{*}{Method} & \multicolumn{3}{c|}{\emph{Pitch}} & \multicolumn{3}{c|}{\emph{Yaw}} & \multicolumn{3}{c|}{\emph{Lighting}} & \multicolumn{3}{c}{\emph{Mixture}} \\
        \cline{3-14}
          & & Arc. & Cos. & Fac. & Arc. & Cos. & Fac. & Arc. & Cos. & Fac. & Arc. & Cos. & Fac. \\
         \hline
         \multirow{5}{*}{{\rotatebox[origin=c]{90}{ArcFace}}}& MIM & 75.65$^*$ & 8.97 & 7.84 & 89.63$^*$ & 11.10 & 8.00 & 94.81$^*$ & 11.33 & 5.76 & 48.45$^*$ & 3.05 & 5.65\\
         & EOT & 86.58$^*$ & 16.16 & 17.48 & 99.63$^*$ & 17.53 & 16.83 & 99.29$^*$ & 17.67 & 12.62 & 73.73$^*$ & 6.78 & 12.46\\
         & GenAP &86.39$^*$ & 27.87 & 31.68 & 99.03$^*$ & 37.80 & 31.17 & 99.33$^*$ & 41.10 & 29.19 &
         68.68$^*$ & 14.89 & 27.18\\
         & Ours($\bm{x}$) & \textbf{94.42}$^*$ & 17.23 & 17.65 & 99.63$^*$ & 21.33 & 17.03 & 99.29$^*$ & 22.86 & 16.81 & 80.88$^*$ & 7.73 & 15.14\\
         & Ours($\bm{w}$) & 94.39$^*$ & \textbf{32.29} & \textbf{31.81} & \textbf{99.90}$^*$ & \textbf{42.27} & \textbf{32.00} & \textbf{99.95}$^*$ & \textbf{47.10} & \textbf{31.33} & \textbf{84.08}$^*$ & \textbf{19.69} & \textbf{31.75}\\
         \hline
        \multirow{5}{*}{\rotatebox[origin=c]{90}{CosFace}}& MIM & 9.61 & 61.32$^*$ & 13.94 & 12.60 & 83.8$^*$ & 13.97 & 13.48 & 95.52$^*$ & 11.71 & 5.24 & 31.08$^*$ & 11.31\\
        & EOT & 21.71 & 75.71$^*$ & 29.45 & 27.77 & 97.53$^*$ & 32.40 & 28.38 & 96.19$^*$ & 28.19 &14.20 & 59.71$^*$ & 26.84
        \\
         & GenAP & 28.74 & 69.45$^*$ & 36.90 & 37.67 & 94.73$^*$ & 36.87 & 40.38 & 98.14$^*$ &34.38 & 20.95 & 48.34$^*$ & 32.72\\
         & Ours($\bm{x}$)& 23.23 & 85.16$^*$ & 29.55 & 28.90 & 97.83$^*$ & 32.80 & 30.71 & 96.90$^*$ & 28.76& 15.18 & 71.03$^*$ & 28.37\\
         & Ours($\bm{w}$) & \textbf{40.06} & \textbf{87.19}$^*$ & \textbf{46.65} & \textbf{51.40} & \textbf{98.13}$^*$ & \textbf{46.20} & \textbf{54.00} & \textbf{98.62}$^*$ & \textbf{45.52} &\textbf{31.30} & \textbf{72.15}$^*$ & \textbf{43.03}\\        
        \hline
        \multirow{5}{*}{\rotatebox[origin=c]{90}{FaceNet}}& MIM & 3.77 & 7.42 & 66.81$^*$ & 7.50 & 10.13 & 67.50$^*$ & 5.00 & 10.29 & 70.52$^*$& 2.34 & 2.88 & 34.80$^*$\\
         & EOT & 9.87 & 17.97 & \textbf{98.10}$^*$ & 13.53 & 20.73 & \textbf{98.87}$^*$ & 12.62 & 22.86 & 96.14$^*$ & 5.70 & 8.19 & 86.73$^*$\\
         & GenAP & 22.35 & 23.29 & 89.61$^*$ & 29.63 & 31.67 & 94.47$^*$ & 26.33 & 31.90 & 92.57$^*$ & 14.56 & 12.77 & 78.72$^*$\\
         & Ours($\bm{x}$) & 15.13 & 21.55 & 98.19$^*$ & 20.27 & 26.67 & 98.57$^*$ & 20.62 & 33.43 & \textbf{98.95}$^*$ &9.56 & 12.59 & 95.94$^*$\\
         & Ours($\bm{w}$) & \textbf{28.94} & \textbf{33.23} & \textbf{98.10}$^*$ & \textbf{38.17} & \textbf{44.47} & 98.70$^*$ & \textbf{38.29} & \textbf{46.48} & 98.29$^*$ &\textbf{21.54} & \textbf{20.54} & \textbf{96.45}$^*$\\
         \hline
    \end{tabular}
    \end{center}
    
    \vspace{-4ex}
\end{table*}

\textbf{Compared methods.}
We compare with \textbf{MIM}~\cite{Dong2017} that integrates a momentum for improving the transferability of adversarial examples, \textbf{EOT}~\cite{Athalye2017Synthesizing} that synthesizes examples over a distribution of transformations, and \textbf{GenAP}~\cite{xiao2021improving} that is a state-of-the-art transferable adversarial method on face recognition based on generative models. We also take \textbf{AdvHat}~\cite{komkov2021advhat} as another baseline by wearing hats, which is also blended into EOT~\cite{Athalye2017Synthesizing} to boost the black-box transferability.

\textbf{Attack types.} We consider three types of physically realizable attacks in the simulation environment, i.e., \textbf{Eyeglass}~\cite{xiao2021improving}, \textbf{Respirator}~\cite{tong2021facesec}, and \textbf{Hat}~\cite{komkov2021advhat} in 3D pasting ways. Then, we mainly adopt Eyeglass for evaluating the vulnerability of face recognition system in the physical world due to its overall excellent black-box performance, which is also consistently observed in~\cite{xiao2021improving}. Besides, we further verify the better practicality of the attack mechanism of 3D
Eyeglass than 2D ones w.r.t. imperceptibility, which can steadily pass defensive mechanism (\emph{commercial Live Detection API}) as stated in Sec.~\ref{sec:phy_exps}. 

\textbf{Implementation details.} We mainly perform impersonation attacks based on the pairs with different identities in this paper. Impersonation attacks are considered more difficult and practical than dodging attacks. The hyperparameters of attack methods and the evaluation results of dodging attacks are presented in Appendix {\color{red} B}.  

\begin{table*}[t]
    \begin{center}
    \scriptsize
    \caption{The attack success rates ($\%$) of the different models against impersonation attacks on CelebA-HQ with adversarial glasses. $^{*}$ indicates white-box attacks.}
    \label{tab:transfer-casia}
    \setlength{\tabcolsep}{0.5pt}
    \begin{tabular}{p{4ex}<{\centering}|c|ccc|ccc|ccc|ccc}
    \hline
      & \multirow{2}{*}{Method} & \multicolumn{3}{c|}{\emph{Pitch}} & \multicolumn{3}{c|}{\emph{Yaw}} & \multicolumn{3}{c|}{\emph{Lighting}} & \multicolumn{3}{c}{\emph{Mixture}} \\
        \cline{3-14}
         & & Arc. & Cos. & Fac. & Arc. & Cos. & Fac. & Arc. & Cos. & Fac. & Arc. & Cos. & Fac. \\
         \hline
         \multirow{5}{*}{{\rotatebox[origin=c]{90}{ArcFace}}}& MIM &  72.71$^*$ & 9.00 & 11.68 & 90.63$^*$ & 10.33 & 11.17 & 94.57$^*$ & 13.05 & 9.71 & 45.58$^*$ & 4.26 & 8.55\\
         & EOT & 81.13$^*$ & 12.94 & 16.81 & 97.33$^*$ & 16.93 & 18.23 & 98.57$^*$ & 16.86 & 15.38& 57.92$^*$ & 5.93 & 12.98\\
         & GenAP & 85.90$^*$ & 30.16 & 39.48 & 99.10$^*$ & 40.33 & 40.23 & 99.10$^*$ & 48.48 & 36.81 & 69.05$^*$ & 19.19 & 35.44\\
         & Ours($\bm{x}$) & 92.19$^*$ & 15.32 & 21.23 & 98.77$^*$ & 20.47 & 23.37 & 99.33$^*$ & 25.86 & 22.71 & 79.74$^*$ & 8.76 & 18.58\\
         & Ours($\bm{w}$) & \textbf{93.84}$^*$ & \textbf{34.23} & \textbf{46.35} & \textbf{99.97}$^*$ & \textbf{47.27} & \textbf{48.70} & \textbf{99.71}$^*$ & \textbf{53.71} & \textbf{44.19} & \textbf{83.04}$^*$ & \textbf{24.29} & \textbf{42.15}\\
         \hline
        \multirow{5}{*}{{\rotatebox[origin=c]{90}{CosFace}}}& MIM & 16.10 & 54.48$^*$ & 19.90 & 21.07 & 77.27$^*$ & 22.83 & 21.33 & 89.76$^*$ & 19.86 & 9.58 & 25.94$^*$ & 16.63\\
         & EOT & 19.32 & 60.06$^*$ & 24.97 & 24.30 & 85.23$^*$ & 29.90 & 26.38 & 94.81$^*$ & 23.76 & 11.69 & 31.81$^*$ & 20.32\\
         & GenAP & 44.29 & 68.13$^*$ & 52.03 & 55.67 & 95.07$^*$ & 55.30 & 56.14 & 98.10$^*$ & 51.57 & 32.31 & 48.62$^*$ & 46.12\\
         & Ours($\bm{x}$) & 28.97 & 83.19$^*$ & 36.06 & 37.50 & 95.40$^*$ & 39.53 & 40.57 & 99.05$^*$ & 36.76 & 21.75 & 65.38$^*$ & 32.23\\
         & Ours($\bm{w}$) & \textbf{49.71} & \textbf{83.68}$^*$ & \textbf{55.29} & \textbf{59.40} & \textbf{95.43}$^*$ & \textbf{58.30} & \textbf{60.95} & \textbf{98.86}$^*$ & \textbf{56.29} & \textbf{39.79} & \textbf{67.31}$^*$ & \textbf{52.92}\\        
        \hline
        \multirow{5}{*}{{\rotatebox[origin=c]{90}{FaceNet}}}& MIM & 8.39 & 6.16 & 68.26$^*$ & 9.97 & 7.70 & 70.47$^*$ & 10.95 & 8.71 & 65.14$^*$ & 5.79 & 3.26 & 34.83$^*$\\
        & EOT & 10.17 & 9.77 & 88.55$^*$ & 13.40 & 11.97 & 92.93$^*$ & 15.24 & 13.38 & 87.76$^*$ & 7.07 & 5.14 & 56.93$^*$\\
         & GenAP & 32.23& 25.81 & 94.45$^*$ & 40.77 & 36.67 & 99.13$^*$ & 40.05 & 40.38 & 94.14$^*$ & 23.33 & 15.91 & 82.79$^*$\\
         & Ours($\bm{x}$) & 20.26 & 18.74 & 98.74$^*$ & 26.73 & 23.67& 99.87$^*$ & 30.43 & 31.62 & 99.67$^*$ & 15.38 & 11.95 & 95.86$^*$\\
         & Ours($\bm{w}$) & \textbf{42.29} & \textbf{36.61} & \textbf{99.77}$^*$ & \textbf{54.20} & \textbf{51.57} & \textbf{100.0}$^*$ & \textbf{52.90} & \textbf{56.76} & \textbf{99.76}$^*$ & \textbf{32.44} & \textbf{25.63} & \textbf{98.47}$^*$\\
         \hline
    \end{tabular}
    \end{center}
    
    \vspace{-4ex}
\end{table*}

\begin{figure}[!t]
\begin{center}
\includegraphics[width=0.9\linewidth]{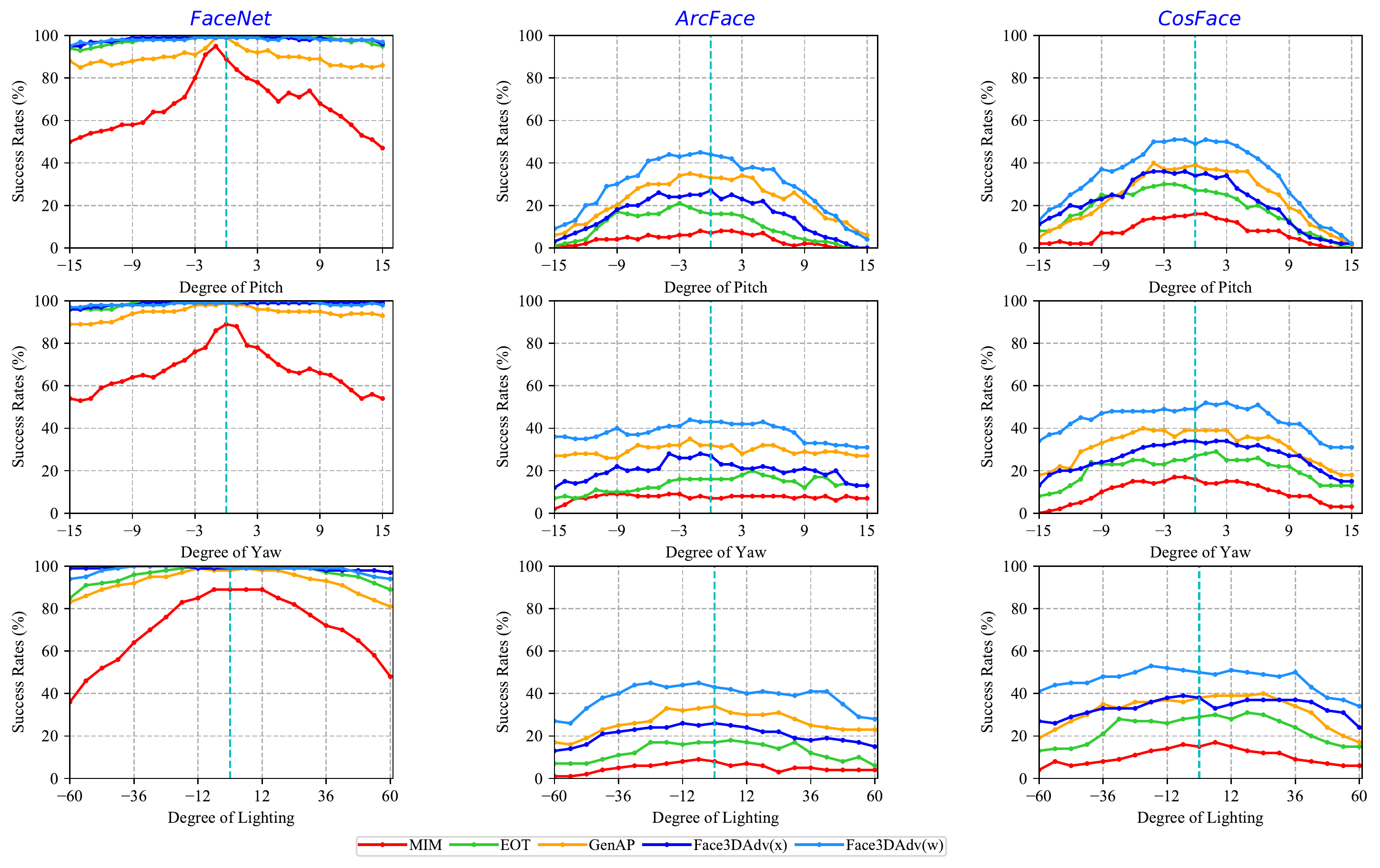}
\end{center}
\vspace{-4ex}
\caption{Attack success rates (\%) of different attacks under various variations, including \emph{pitch}, \emph{yaw}, and \emph{lighting}. FaceNet~\cite{schroff2015facenet} is chosen as a white-box model, and test the performance in the chosen three models.}
\label{fig:curve}
\vspace{-3ex}
\end{figure}

\begin{figure*}[t]
\begin{center}
\vspace{-0.4cm}
\includegraphics[width=0.99\linewidth]{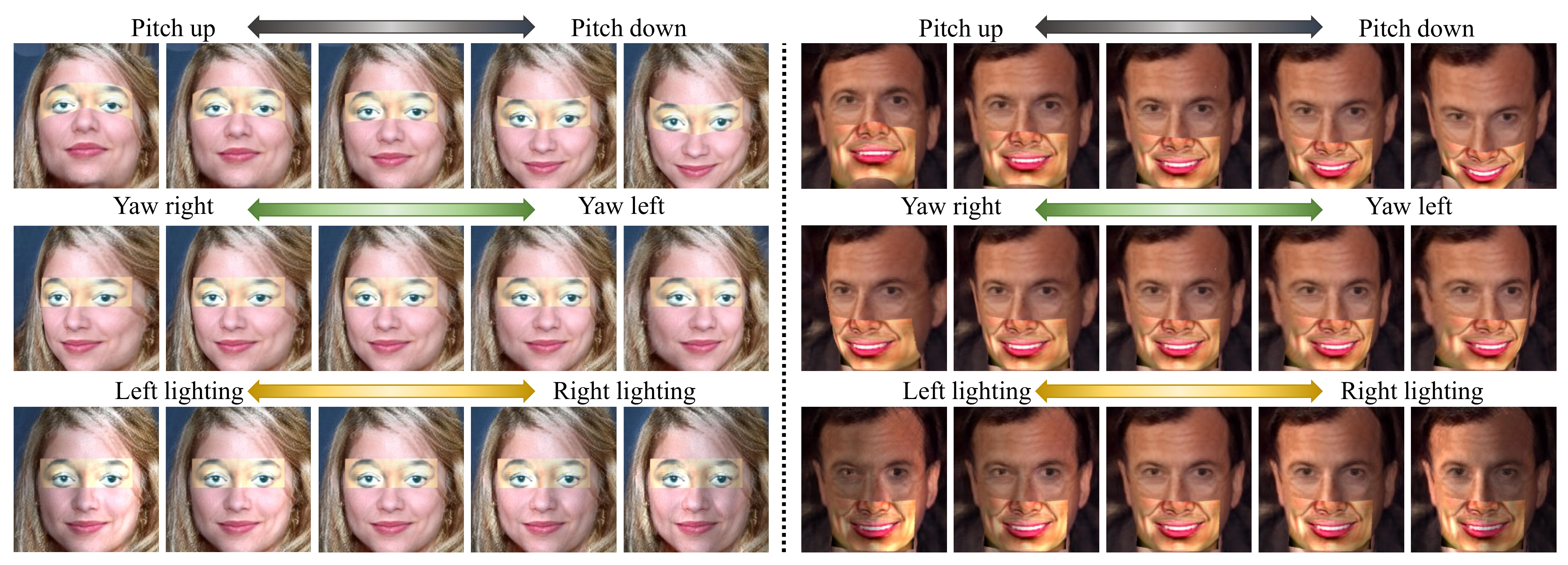}
\end{center}
\vspace{-4ex}
\caption{Sample results in simulation framework for physical attacks of \textbf{Eyeglass} and \textbf{Respirator}, which realizes 3D control of the adversarial examples, including \emph{pitch}, \emph{yaw} and \emph{Lighting}. Thus, the framework can be reliably  used as a surrogate for implementing physical adversarial attacks on face recognition due to cheap and easy implementation. More detailed examples are presented in Appendix {\color{red} C}.}
\vspace{-3ex}
\label{fig:visual}
\end{figure*}

\subsection{Benchmarking on Simulation Framework}
In this section, we compare the performance of different attacks for comprehensive evaluations on face recognition, based on the proposed simulation framework. Fig.~\ref{fig:visual} shows examples for a physical attack in the simulation framework, which effectively achieves 3D control of the adversarial examples. 

\textbf{Effectiveness of the proposed method.}
To verify the effects of different face variations, including \emph{pitch}, \emph{yaw} and \emph{lighting}, we compare the performance of different methods. Table~\ref{tab:transfer} and Table~\ref{tab:transfer-casia} show the attack success rates ($\%$) of the different face recognition models on LFW and CelebA-HQ, respectively. We can see that different face variations weaken the attack performance of the methods in varying degrees, especially for the effect of \emph{mixture} type. Despite this, Face3DAdv with two variations leads to higher white-box attack success of face recognition models. The results also demonstrate that Face3DAdv can achieve more robust and effective testing performance, benefitting from various physical variations in the optimization phase.

\begin{table}[t]
\vspace{-1ex}
    \begin{center}
    \scriptsize
    \caption{Comparison of AdvHat~\cite{komkov2021advhat} and ours by two types of physically realizable attacks. CosFace is a white-box model.}
    \label{tab:compare}
    \setlength{\tabcolsep}{5pt}
    \begin{tabular}{c|c|c|ccc|c}
    \hline
      \multirow{2}{*}{Type} & \multirow{2}{*}{Testing} & \multirow{2}{*}{Method} & \multicolumn{4}{c}{Face variations} \\
       \cline{4-7}
       & & & \emph{Pitch} & \emph{Yaw} & \emph{Lighting} & \emph{Mixture} \\
       \hline
       \multirow{6}{*}{\textbf{Hat}}  & \multirow{2}{*}{ArcFace}& AdvHat~\cite{komkov2021advhat} &   2.97 & 4.37 & 4.43 & 2.34 \\
       && Face3DAdv & 11.13 & 11.83 & 12.24 & 8.63\\
       \cline{2-7}
       & \multirow{2}{*}{CosFace}& AdvHat~\cite{komkov2021advhat} &  63.13 & 84.50 & 89.76 & 39.44\\
       && Face3DAdv & 75.29 & 81.03 & 80.00 & 56.45 \\
       \cline{2-7}
       & \multirow{2}{*}{FaceNet}& AdvHat~\cite{komkov2021advhat} &  3.35 & 3.80 & 4.86 & 4.60\\
       && Face3DAdv & 9.45 & 9.57 & 9.29 & 9.89\\
       \hline
       \multirow{6}{*}{\textbf{Respirator}} & \multirow{2}{*}{ArcFace}& AdvRespirator~\cite{komkov2021advhat} &   19.97 & 24.83 & 26.71 & 12.71 \\
       && Face3DAdv & 36.26 & 48.43 & 49.10 & 29.02\\
       \cline{2-7}
       &\multirow{2}{*}{CosFace}& AdvRespirator~\cite{komkov2021advhat} & 74.77 & 94.83 & 95.67 & 47.44 \\
       & &Face3DAdv &  89.58 & 96.30 & 96.29 & 67.31\\
       \cline{2-7}
       &\multirow{2}{*}{FaceNet}& AdvRespirator~\cite{komkov2021advhat} &  16.45 & 18.53 & 18.67 & 12.47\\
       & &Face3DAdv & 30.65 & 31.67 & 32.95 & 26.52\\
       \hline
    \end{tabular}
    \end{center}
\vspace{-3ex}
\end{table}

\begin{figure}[t]
\begin{center}
\includegraphics[width=0.9\linewidth]{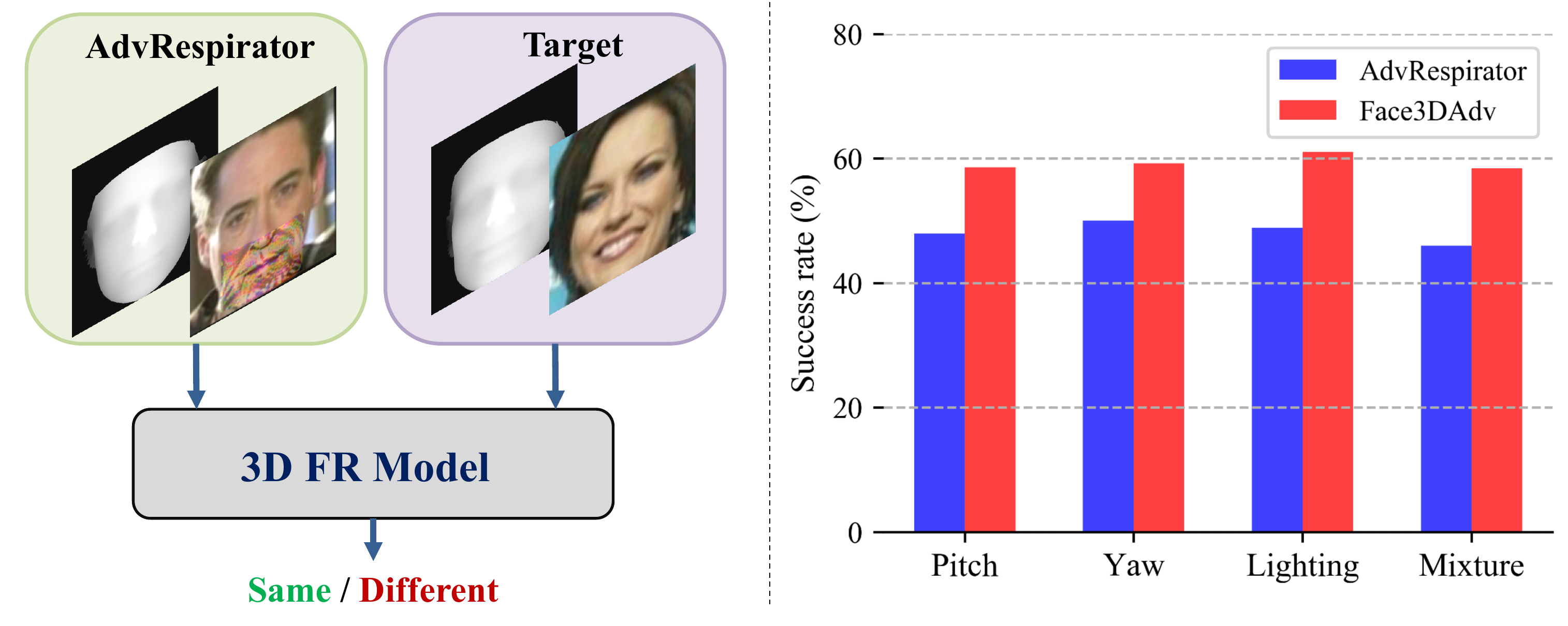}
\end{center}
\vspace{-4ex}
\caption{The attack success rate ($\%$) of two methods against black-box 3D face recognition model by using the attack type of Respirator.}
\label{fig:3dfr}
\vspace{-3ex}
\end{figure}

\begin{table}[t]
    \begin{center}
    \small 
    \caption{Ablation study of the importance sampling strategy. `w/o IS' indicates equally sampling. CosFace is a white-box model.}
    \label{tab:ab}
    \setlength{\tabcolsep}{7pt}
    \begin{tabular}{c|cc|cc|cc}
    \hline
      \multirow{2}{*}{Testing}
      & \multicolumn{2}{c|}{\emph{Pitch}} & \multicolumn{2}{c|}{\emph{Yaw}} & \multicolumn{2}{c}{\emph{Lighting}} \\
      \cline{2-7}
      & w/o IS & with IS & w/o IS & with IS & w/o IS & with IS \\
      \hline
      {ArcFace} & 39.12 & 40.06 & 50.07 & 51.40 & 52.62 & 54.00\\
      {CosFace} & 84.90 & 87.19 & 98.37 & 98.13 & 98.10 & 98.62\\
      {FaceNet}& 45.16 & 46.65 & 46.03 & 46.20 & 45.67 & 45.52\\
      \hline
    \end{tabular}
    \end{center}
    
    \vspace{-3ex}

\end{table}

    


\textbf{Transferability of the proposed method.} We then feed the crafted adversarial images against one face model into other models for testing the transferablity. The results indicate that Face3DAdv can obtain better black-box transferability in the simulation framework. Meanwhile, Fig.~\ref{fig:curve} shows the detailed performance of the different face variations based on white-box FaceNet. Note that Face3DAdv ($\bm{w}$) performs best where the axis of face conditions belongs to zero, revealing that our method can consistently enhance the black-box performance even in \emph{without} variations.

\textbf{Comparison with AdvHat~\cite{komkov2021advhat}.} We compare the performance of our method with AdvHat by adopting the attack type of \emph{Hat}. In addition, we introduce its variation of attack type based on \emph{Respirator}, named as AdvRespirator.
Table~\ref{tab:compare} shows the comparable results in these two physically realizable attacks. We found that  the optimized region of Hat is not very prominent in the whole face region, making it hard to fully utilize the information of 3D variations in the white-box optimization phase. Nevertheless, our method consistently obtains better performance in terms of effectiveness and transferability in these two physically realizable attacks.


\textbf{Effectiveness on 3D face recognition model.} Since the proposed method lies in textured-based attacks almost without changing the depth map of a face, it should be able to attack 3D face recognition by leveraging the black-box transferability. To verify this, we introduce typical RGBD-FR~\cite{xiong2019improving} that utilizes depth images to explore the global facial layout. Fig.~\ref{fig:3dfr} shows the attack success rates against RGBD-FR based on the attack type of Respirator. The results show that our method by texture-based attacks can achieve effective attacks against 3D face recognition based on black-box transferability.

\textbf{Ablation study of the importance sampling strategy.} We conduct an ablation study to investigate the effects of the importance sampling strategy introduced in Sec.~\ref{sec:is}. Table~\ref{tab:ab} shows the attack success rates \emph{with} and \emph{without} importance sampling. After introducing this strategy, ours can better exploit profitable transformations in the optimizing phase, making it more effective during the testing phase.


\begin{figure}[t]
\begin{center}
\includegraphics[width=0.8\linewidth]{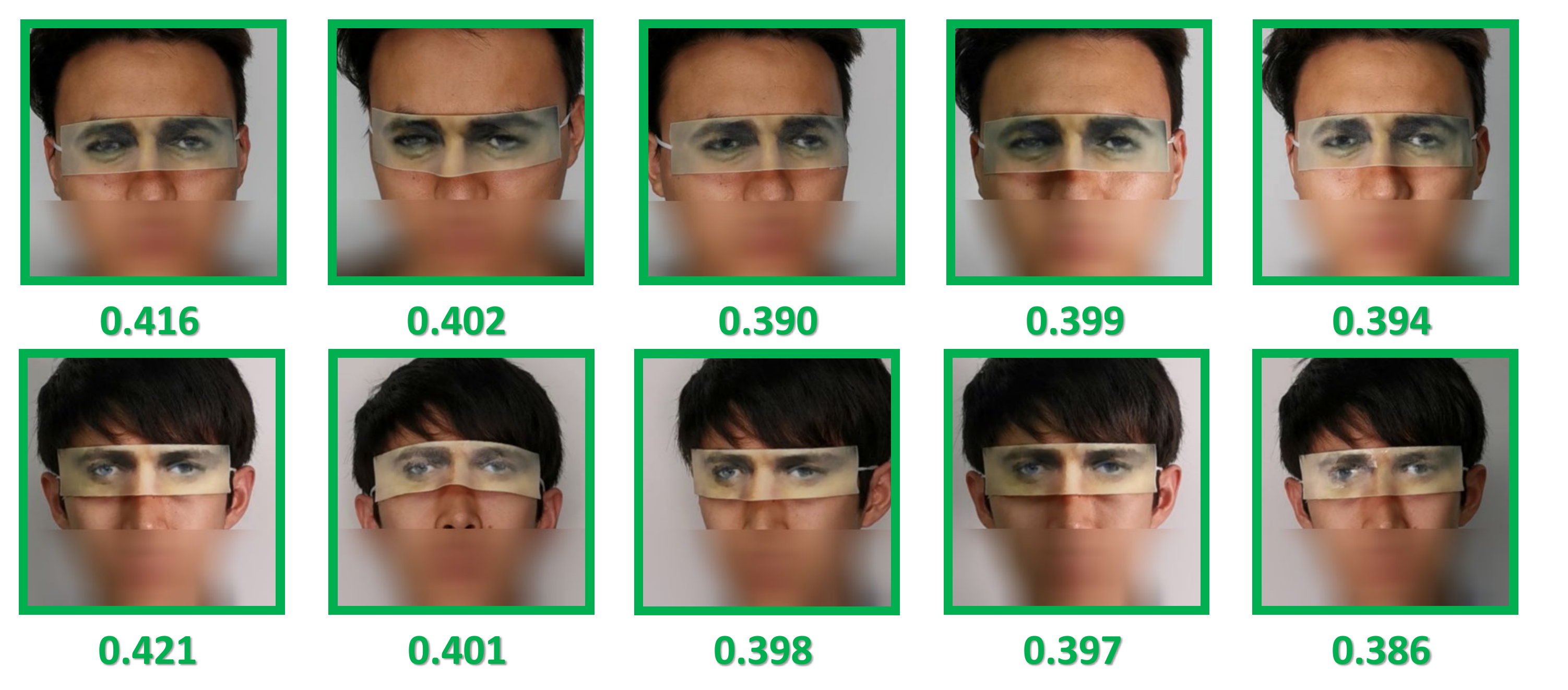}
\end{center}
\vspace{-4ex}
\caption{Experimental results of physical attacks by wearing the 3D adversarial glasses. Green box refers to successful physical attacks, and similarity scores are marked.}
\label{fig:phy}
\vspace{-3ex}
\end{figure}
\begin{table}[t]
    \begin{center}
    \small 
    \caption{The attack success rates (\%) of physical experiments with 3D adversarial glasses against CosFace. We also adopt Live Detection API to test the imperceptibility.}
    \label{tab:phy}
    \setlength{\tabcolsep}{6pt}
    \begin{tabular}{c|c|ccc|c}
    \hline
      \multirow{2}{*}{Environment} 
       & \multirow{2}{*}{Method} & \multicolumn{3}{c|}{\textbf{Effectiveness}} & {\textbf{Imperceptibility}} \\
       \cline{3-6}
       & & \emph{Pitch} & \emph{Yaw} & \emph{Lighting} & \emph{Live Detection} \\
       \hline
       \multirow{2}{*}{Physical} & GenAP & 42.43 & 53.50& 57.70 & 5.11 \\
        & Face3DAdv & 64.62&69.37 & 72.40 & 85.52\\       
       \hline
    \end{tabular}
    \end{center}
    \vspace{-3ex}

\end{table}

\vspace{-2ex}
\subsection{Experiments in the Physical World}
\label{sec:phy_exps}
\vspace{-0.0cm}
In this section, we aim to verify that the simulator can approximate the performance of attacks in the physical world. Therefore, we invited five volunteers (the consent was obtained and visual images were encrypted) to be the attackers, and assigned an identity from LFW as the victim for this experiment. The main steps were as follows: First, we took a face photo of a volunteer with a fixed camera under natural light.
Then, we used the simulation framework for adversarial attacks under different variations and get an adversarial glasses for each volunteer. The adversarial glasses were 3D-printed and pasted on real faces. Finally, after wearing adversarial glasses, the volunteers tried to reproduce different conditions, including some specific poses and lighting via a stabilized environment source. Fig.~\ref{fig:phy} and Table~\ref{tab:phy} illustrates the effectiveness of our method under varying face variations over the baseline in the real world, which is already seen in the simulation framework. The main reason is because Face3DAdv benefits from various simulated physical transformations, and present the consistent performance in the real world. In addition, 3D texture-based attack is also more conducive to passing commercial \textbf{Live Detection API} steadily since 3D texture-based attack does not almost change the depth. We also provide more details in Appendix {\color{red} D}.

\section{Conclusion}
In this paper, we introduce a simulation framework based on 3D face modeling, which can control different face variations and physical conditions to conduct reproducible evaluations. Based on this, we also propose Face3DAdv to craft more robust adversarial patches by considering the 3D face transformations. Extensive experimental results verify the consistent improvements over the previous methods in both simulated and physical environments, against white-box and black-box face recognition models.


%
%
\bibliographystyle{splncs04}
\bibliography{egbib}

\clearpage
\appendix

\section{Detailed Testing Protocol}

In the simulation framework, we choose
a total of 200 experimenters from LFW and CelebA-HQ, which are near-front angles. For every experimenter, we introduce a controllable environmental testing protocol including different poses and lightings as follows.

\begin{itemize}
    \item[1)] \textbf{Pitch:} based on the proposed simulation framework, we control specific movement ranges of the cruciform rail from $-15$ to $15$ angles, and evaluate the performance of attack methods by using the obtained image of every angle. Thus there are a total of $30$ images for every experimenter.
    \item[2)] \textbf{Yaw:} we similarly control movement ranges of the cruciform rail from $-15$ to $15$ angles, and evaluate the performance of attack methods for images for every angle. Therefore, there are a total of $30$ images for every experimenter in this type.
    \item[3)] \textbf{Lighting:} we obtain relighted testing images by creating a shading map of lighting from $-60$ to $60$ degrees. There are a total of 20 images for every experimenter in this type when the sampling interval is set to $6$.
    \item[4)] \textbf{Mixture:} We have linearly combined these three conditions to constitute a new type, named mixture. Specifically, we sample uniformly at intervals of $6$ under $[-15, 15]$ degrees of yaw and pitch, respectively, meanwhile setting three different degrees of lighting as -40, 0 and 40. Thus there are a total of 108 images for every experimenter in this type.
\end{itemize}

In total, our testing protocol in the simulation framework consists of 200 experimenters and a total of 37,600 testing faces. Therefore, a wide range of different
physical types in the evaluation, far ahead of the previous datasets, makes our testing protocol challenging and realistic for the existing attack methods. 

\textbf{Evaluation of 2D transformations.} We consider three types of 2D physical transformations, which are rotation, projective transformation and their mixture as follows.

\begin{itemize}
    \item[1)] \textbf{Rotation:} the angle of the rotation is sampled from $\mathcal{N}(0, \sigma_{1})$.
    \item[2)]\textbf{Projective transformation:} it has eight parameters including $[a_{0}, a_{1}, a_{2}, b_{0},$ $b_{1}, b_{2},c_{0}, c_{1}]$. Given a point $(x, y)$, we can calculate the mapping point $(x', y') = ((a_{0}x+a_{1}y+a_{2})/k, (b_{0}x+b_{1}y+b_{2})/k,)$, where $k = c_{0}x + c_{1}y + 1$. $a_{0}$ and $b_{0}$ are sampled from $\mathcal{N}(1, \sigma_{1})$, and other parameters are sampled from $\mathcal{N}(0, \sigma_{1})$.
    \item[3)]\textbf{Mixture-2D:} We orderly combine these two conditions to constitute a  new type, named Mixture-2D.
    
\end{itemize}

In the evaluation of 2D transformation, we set the fixed random seed and sample $\sigma$ uniformly from $\mathcal{U}(0, 0.1)$. Table~\ref{tab:compare-2d} shows comparison of EOT and Face3DAdv by 2D variation types. We can see that the performance of white-box attack between the two methods is close to 100\%, indicating that the methods can resist the effect of 2D variations in certain varying degrees. The main reason is that the 2D variations can be easily integrated into the optimization phase. Meanwhile, Face3DAdv can obtain better black-box transferability due to the involvement of various 3D physical conditions. Therefore, 3D transformations can be regarded as more difficult and practical than 2D transformations, which also further encourages us to evaluate the performance of different attack methods in varying 3D physical transformations.

\begin{table}[t]
    \begin{center}
    \small 
    \caption{Comparison of EOT and ours by 2D variation types. CosFace is a white-box model.}
    \label{tab:compare-2d}
    \setlength{\tabcolsep}{3pt}
    \begin{tabular}{c|c|ccc}
    \hline
       \multirow{2}{*}{Testing} & \multirow{2}{*}{Method} & \multicolumn{3}{c}{Face variations} \\
       \cline{3-5}
       & & \emph{Rotation} & \emph{Projection} & \emph{Mixture-2D} \\
       
       \hline
       \multirow{2}{*}{ArcFace}& EOT &  16.0 & 10.0 & 12.0 \\
       & Face3DAdv & 18.0 & 12.0 & 16.0 \\
       \hline
       \multirow{2}{*}{CosFace}& EOT & 99.0 & 99.0 & 99.0 \\
       & Face3DAdv & 98.0 & 98.0 & 98.0\\
       \hline
       \multirow{2}{*}{FaceNet}& EOT &  13.0 & 13.0 & 13.0 \\
       & Face3DAdv & 18.0 & 18.0 & 18.0\\

       \hline
    \end{tabular}
    \end{center}
\end{table}

\begin{table*}[t]
    \begin{center}
    \scriptsize
    \caption{The attack success rates ($\%$) of the different face recognition models against dodging attacks on LFW with adversarial glasses. $^{*}$ indicates white-box attacks.}
    \label{tab:transfer-dodging}
    \setlength{\tabcolsep}{0.5pt}
    \begin{tabular}{p{4ex}<{\centering}|c|ccc|ccc|ccc|ccc}
    \hline
         & \multirow{2}{*}{Method} & \multicolumn{3}{c|}{\emph{Pitch}} & \multicolumn{3}{c|}{\emph{Yaw}} & \multicolumn{3}{c||}{\emph{Lighting}} & \multicolumn{3}{c}{\emph{Mixture}} \\
        \cline{3-14}
          & & Arc. & Cos. & Fac. & Arc. & Cos. & Fac. & Arc. & Cos. & Fac. & Arc. & Cos. & Fac. \\
         
         \hline
         \multirow{5}{*}{\rotatebox[origin=c]{90}{ArcFace}}& MIM & \textbf{100.0}$^*$ & 65.19 & 65.19 & 99.93$^*$ & 60.23 & 62.17 & 100.0$^*$ & 69.81 & 70.81 & 99.79$^*$ & 85.10 & 75.32\\
         & EOT & \textbf{100.0}$^*$ & 76.32 & 76.61 & \textbf{100.0}$^*$ & 74.40 & 74.43 & \textbf{100.0}$^*$ & 82.57 & 81.67 & 99.98$^*$ & 90.10 & 85.67\\

         & GenAP & \textbf{100.0}$^*$ & 95.29 & 97.45 & \textbf{100.0}$^*$ & 96.23 & 97.73 & \textbf{100.0}$^*$ & 97.43 & 97.81 & \textbf{100.0}$^*$ & 98.44 & 98.38\\
         & Ours($\bm{x}$) & \textbf{100.0}$^*$ & 86.94 & 87.32 & \textbf{100.0}$^*$ & 86.20 & 85.93 & \textbf{100.0}$^*$ & 90.00 & 87.48 & \textbf{100.0}$^*$ & 94.87 & 91.23\\
         & Ours($\bm{w}$) & \textbf{100.0}$^*$ & \textbf{96.52} & \textbf{98.39} & \textbf{100.0}$^*$ & \textbf{96.33} & \textbf{98.13} & \textbf{100.0}$^*$ & \textbf{97.69} & \textbf{97.97} & \textbf{100.0}$^*$ & \textbf{98.91} & \textbf{98.77}\\
         \hline
        \multirow{5}{*}{\rotatebox[origin=c]{90}{CosFace}}& MIM & 45.03 & 99.77$^*$ & 57.32 & 27.37 & \textbf{100.0}$^*$ & 57.10 & 39.71 & \textbf{100.0}$^*$ & 63.76 & 68.32 & 99.31$^*$ & 70.08\\
         & EOT & 49.39 & 99.97$^*$ & 61.13 & 33.20 & \textbf{100.0}$^*$ & 60.87 & 45.81 & \textbf{100.0}$^*$ & 64.71 & 71.78 & 99.78$^*$ & 73.66\\
         & GenAP & 86.42 & 99.94$^*$ & 98.29 & 81.27 & \textbf{100.0}$^*$ & 97.90 & 85.71 & \textbf{100.0}$^*$ & 97.71 & 93.87 & 99.99$^*$ & \textbf{98.94}\\
         & Ours($\bm{x}$) & 67.77 & \textbf{100.0}$^*$ & 88.81 & 54.63 & \textbf{100.0}$^*$ & 90.53 & 68.62 & \textbf{100.0}$^*$ & 91.29 & 84.34 & \textbf{100.0}$^*$ & 94.23\\
         & Ours($\bm{w}$) & \textbf{89.32} & \textbf{100.0}$^*$ & \textbf{98.65} & \textbf{85.80} & \textbf{100.0}$^*$ & \textbf{98.33} & \textbf{87.76} & \textbf{100.0}$^*$ & \textbf{98.19} & \textbf{95.42} & \textbf{100.0}$^*$ & 98.79\\        
        \hline
        \multirow{5}{*}{\rotatebox[origin=c]{90}{FaceNet}}& MIM & 45.84 & 61.71 & 98.52$^*$ & 29.07 & 57.37 & 97.60$^*$ & 39.38 & 62.14 & 99.76$^*$ & 67.92 & 83.03 & 97.56$^*$\\
        & EOT & 52.00 & 78.26 & \textbf{100.0}$^*$ & 37.07 & 76.77 & \textbf{100.0}$^*$ &47.10 & 79.81 & \textbf{100.0}$^*$ & 73.36 & 91.01 & \textbf{100.0}$^*$\\
         & GenAP & 92.42 & 96.03 & \textbf{100.0}$^*$ & 88.83 & 98.13 & \textbf{100.0}$^*$ & 90.05 & 98.38 & \textbf{100.0}$^*$ & 96.31 & 98.38 & \textbf{100.0}$^*$\\
         & Ours($\bm{x}$) & 66.74 & 93.84& \textbf{100.0}$^*$ & 52.20 & 92.80 & \textbf{100.0}$^*$ & 67.90 & 94.10 & \textbf{100.0}$^*$ & 82.47 & 97.06 & \textbf{100.0}$^*$\\
         & Ours($\bm{w}$) & \textbf{96.06} & \textbf{98.71} & \textbf{100.0}$^*$ & \textbf{96.23} & \textbf{99.03} & \textbf{100.0}$^*$ & \textbf{97.10} & \textbf{99.00} & \textbf{100.0}$^*$ & \textbf{98.17} & \textbf{99.60} & \textbf{100.0}$^*$\\
         \hline
    \end{tabular}
    \end{center}
    \vspace{-2ex}
\end{table*}


\begin{figure*}[t]
\begin{center}
\includegraphics[width=0.99\linewidth]{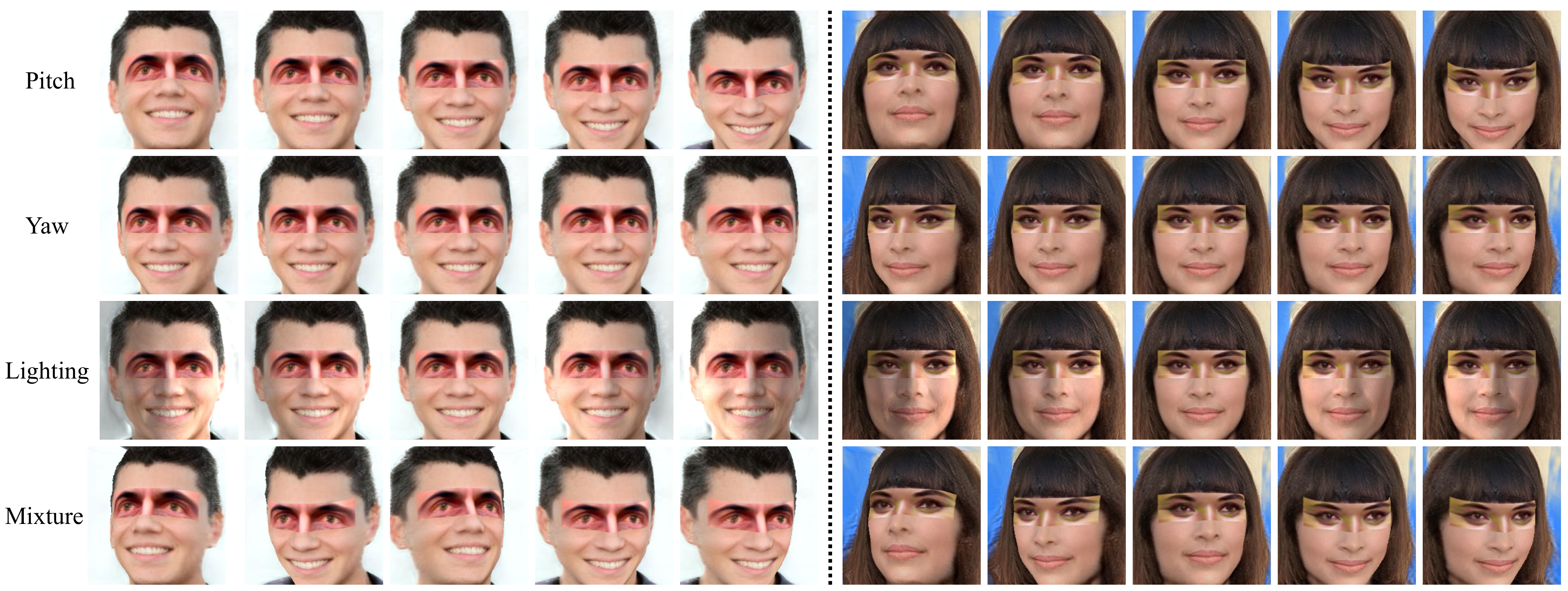}
\end{center}
\vspace{-2ex}
\caption{Sample results in simulation framework for physical attack on CelebA-HQ, which realizes 3D control of the adversarial examples, including \emph{pitch}, \emph{yaw}, \emph{lighting} and \emph{mixture}.}
\vspace{-1ex}
\label{fig:exps1}

\begin{center}
\includegraphics[width=0.99\linewidth]{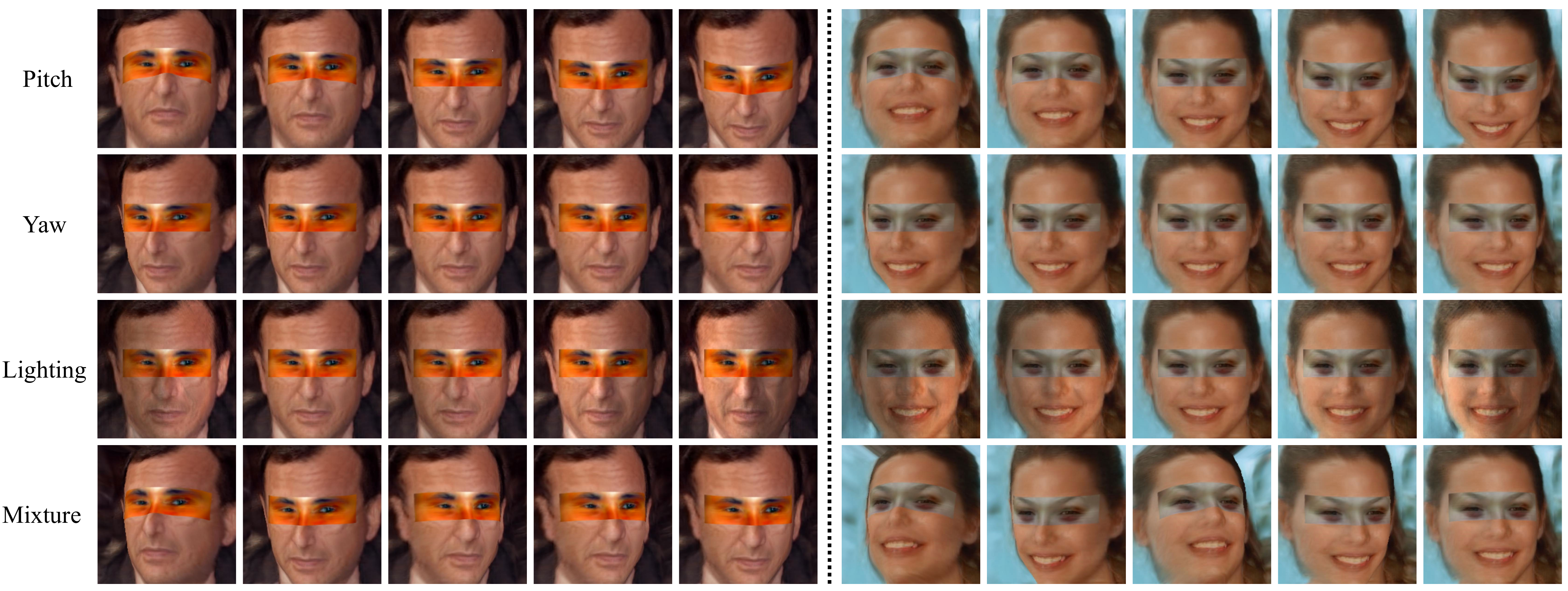}
\end{center}
\vspace{-2ex}
\caption{Sample results in simulation framework for physical attack on LFW, which realizes 3D control of the adversarial examples, including \emph{pitch}, \emph{yaw}, \emph{lighting} and \emph{mixture}.}
\vspace{-1ex}
\label{fig:exps2}

\begin{center}
\includegraphics[width=0.99\linewidth]{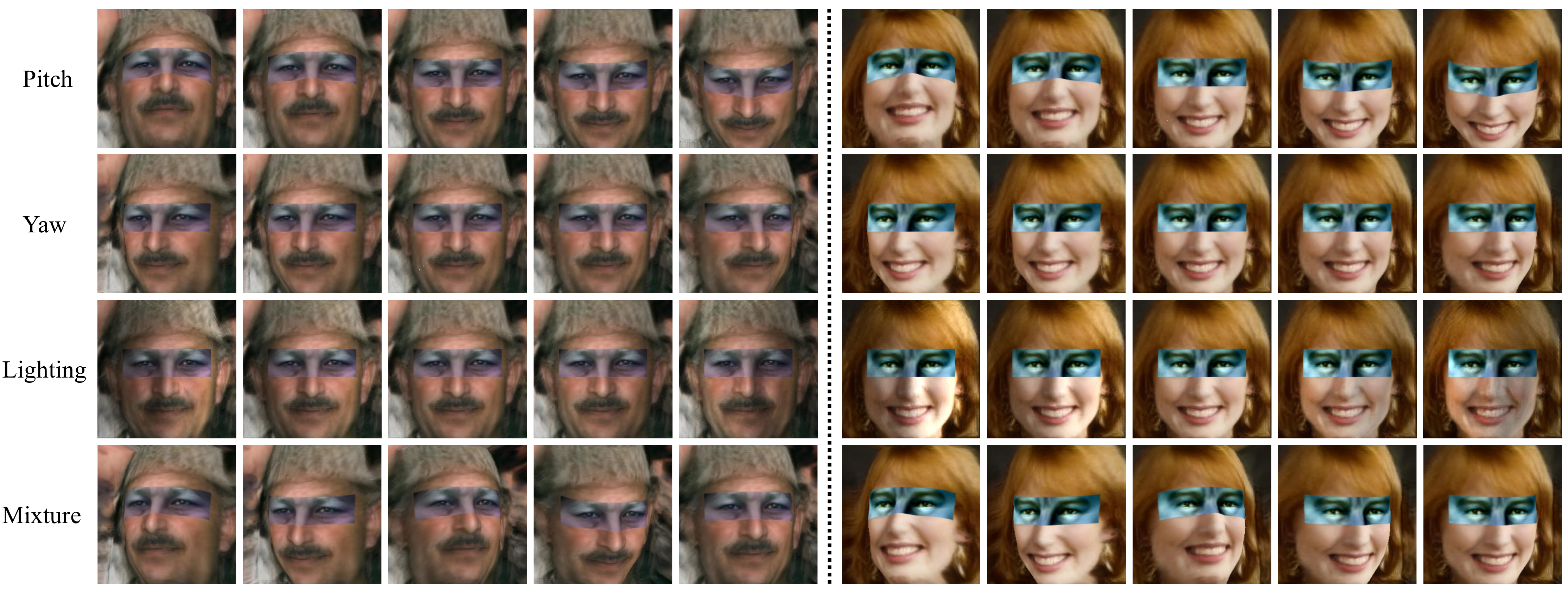}
\end{center}
\vspace{-2ex}
\caption{Sample results in simulation framework for physical attack on LFW, which realizes 3D control of the adversarial examples, including \emph{pitch}, \emph{yaw}, \emph{lighting} and \emph{mixture}.}
\vspace{-1ex}
\label{fig:exps3}
\end{figure*}

\section{More Experiments}

\subsection{Implementation Details}

Note that MIM and EOT select optimal parameters as report for black-box performance  by following~\cite{xiao2021improving}. We thus set the number of iterations as $N = 400$, the learning rate $\alpha = 1.5$, the decay factor $\mu = 1$, and the size of perturbation $\epsilon = 40$ for impersonation and $\epsilon = 255$ for dodging  under the $\ell_{\infty}$ norm bound, which are identical for all the experiments. The sampling number of EOT is set as $M = 10$. And GenAP adopts original public hyperparameters. As for Face3DAdv, We set the number of iterations $N_1 = 300$, $N_2=100$, and the learning rate of Adam optimizer $\eta = 0.01$. Besides, we sample $10$ transformations from $20$ candidates for Ours in every optimization step.

\subsection{Training Efficiency}
We set the number of iterations as $N$ and the sampling number of EOT as $M$ in baselines, thus the adversarial patch is generated by $N * M$ forward and backward propagations. As a comparison, our method requires sampling $M$ times from $M_{L}$ candidates ($M_{L} = 2 * M$ in our setting) at every iteration, thus needs to perform $N * M_{L}$ forward propagations and $N * M$ backward propagations. Overall, we only use acceptable overhead on running complexity in the inference phase, and obtain a better performance.

\subsection{Evaluation of Dodging Attacks}
We perform dodging attacks based
on the pairs of images with the same identities on LFW. Table~\ref{tab:transfer-dodging} shows the attack success rates (\%) of the different face recognition models against dodging attacks on LFW with adversarial glasses. We can see that the overall success rates of dodging attacks are very high, which illustrate that impersonation attacks are more difficult than dodging attacks. Despite this, Face3DAdv with two variations leads to higher white-box and black-box success rates of face recognition models. Similar to the conclusion in impersonation attacks, the results of dodging attacks also demonstrate that Face3DAdv can achieve more robust and effective testing performance. The main reason is that Face3DAdv benefits from various physical variations in the optimization phase.

\section{More Examples}


In Fig.~\ref{fig:exps1}, Fig.~\ref{fig:exps2} and Fig.~\ref{fig:exps3}, we show more results of Eyeglass in simulation framework for physical attack on different datasets, which effectively realize 3D control of the adversarial examples, including pitch, yaw, lighting, and mixture. Thus, the framework can be reliably used as a surrogate for implementing physical adversarial attacks on face recognition due to cheap and easy implementation.

\section{Physical Evaluations}

In physical experiments, we mainly present the following steps. First, we took a face photo of a volunteer with a fixed camera under natural light. Then, we used the simulation framework for adversarial attacks under different variations and get  adversarial glasses for each volunteer. The adversarial glasses were 3D-printed and pasted on real faces. Finally, after wearing adversarial glasses, the volunteers tried to reproduce different conditions, including some specific yaw, pitch and lighting via a stabilized environment source. 


\textbf{Live detection.} To demonstrate the effectiveness of Face3DAdv in live detection, we choose a powerful commercial live detection API service. The working mechanism and training data are completely unknown for us. We then feed the crafted adversarial images into the black-box API for evaluating the effectiveness. We obtain a satisfying performance on passing the live detection API with a success rate of 85.52\% under diverse variations, which outperforms 2D methods by a
margin. Since 3D texture-based attack does not almost change the depth map of a face, it is also more conducive to passing commercial Live Detection API
steadily.

\end{document}